\documentclass[12pt]{article}

\usepackage{setspace}

\usepackage{color}   
\usepackage{hyperref}
\hypersetup{
	colorlinks=true, 
	linktoc=all,     
	linkcolor=blue,  
	citecolor=red,
}
\usepackage{geometry}                
\usepackage{graphicx}
\usepackage{amssymb}
\usepackage{amsmath}
\usepackage{appendix}
\usepackage{tikz}
\usepackage[numbers]{natbib}

\usepackage{booktabs}       
\usepackage{multirow}

\usepackage{scrextend}

\usetikzlibrary{positioning}
\usetikzlibrary{arrows}

\newtheorem{theorem}{Theorem}[section]

\numberwithin{figure}{section}
\numberwithin{equation}{section}

\usepackage{algorithm}
\usepackage{algorithmic}
\usepackage{bm}

\usepackage{stmaryrd}

\begin{document}
	
	\title{An Interpretive Constrained Linear Model for \\ResNet and MgNet}
	\author{Juncai He\footnotemark[1] \quad Jinchao Xu\footnotemark[2]\quad  Lian Zhang\footnotemark[2] \quad Jianqing Zhu\footnotemark[3]}
	\date{} 
	
	\maketitle
	
	\renewcommand{\thefootnote}{\fnsymbol{footnote}} 
	\footnotetext[1]{Department of Mathematics, The University of Texas at Austin, Austin, TX 78712, USA (juncai.he@kaust.edu.sa).} 
	\footnotetext[2]{Department of Mathematics, The Pennsylvania State University, University Park, PA 16802, USA (xu@math.psu.edu, luz244@psu.edu).} 
	\footnotetext[3]{Faculty of Science, Beijing University of Technology, Beijing 100124, China (jqzhu@emails.bjut. edu.cn).} 
	
	\begin{abstract}
		We propose a constrained linear data-feature-mapping model 
		as an interpretable mathematical model for image classification using a 
		convolutional neural network (CNN). 
		From this viewpoint, we establish detailed connections 
		between the traditional iterative schemes for linear systems 
		and the architectures of the basic blocks of ResNet- and MgNet-type models. 
		Using these connections, we present some modified ResNet models that 
		compared with the original models have fewer parameters and yet can produce
		more accurate results, thereby demonstrating the validity of this constrained learning data-feature-mapping assumption. 
		Based on this assumption, we further propose a general data-feature iterative scheme to show the rationality of MgNet. We also provide a systematic numerical study on MgNet to show its success and advantages in image classification problems and demonstrate its advantages in comparison with established networks.
	\end{abstract}


\section{Introduction}
This paper focuses on providing mathematical insight into deep
convolutional neural network (CNN) models that have been successfully
applied in many machine learning and artificial intelligence areas
such as computer vision, natural language processing, and reinforcement
learning~\cite{lecun2015deep}.
Examples of CNN models include the LeNet-5 model presented by LeCun et al. in
1998~\cite{lecun1998gradient}, the AlexNet model by Hinton et al. in 2012
\cite{krizhevsky2012imagenet}, the residual network (ResNet) by  He et
al. in 2015~\cite{he2016deep}, pre-act ResNet in
2016~\cite{he2016identity}, MgNet in 2019~\cite{he2019mgnet}, and other variants of CNN \cite{simonyan2014very, szegedy2015going, huang2017densely}.  
Among these CNN models, ResNet and pre-act ResNet are of
special theoretical and practical interest.  
In fact, researchers have taken many steps to advance the field's theoretical understanding of ResNet, to explain how and why it works well, and to design better residual-type architecture based on empirical observations and informal
interpretation \cite{zagoruyko2016wide, larsson2016fractalnet,
	gomez2017reversible, xie2017aggregated, zhang2017polynet,
	szegedy2017inception, huang2017densely}.
Most of the works mentioned here focus on the interpretation of the basic block in ResNet. However, some fine structures of ResNet and MgNet remain unclear. For example, how do we interpret the convolutional kernels, what is the role of the activation function, and how do we explain the pooling operations? For the explanation of pooling operations in particular, as far as we are aware, almost no literature provides deep insights from a mathematical viewpoint. Given the natural multi-scale (multi-resolution) structure and the residual correction iterative scheme in multigrid methods~\cite{xu1992iterative,hackbusch2013multi,xu2017algebraic}, we were inspired and motivated to interpret MgNet and ResNet architecture (the whole feature extraction process including pooling layers) from the multigrid and iterative methods perspectives.

We propose a generic mathematical model underlying the
basic blocks of ResNet and MgNet to demonstrate their dual relation and 
understand how they function.
At the core of our model is the following assumption:  there is a data-feature 
mapping
\begin{equation}\label{Auf}
	A\ast u = f,
\end{equation}
where $A$ is understood as the underlying data-feature mapping to be learned and in practice is implemented as a conversational kernel with multi-channel. In addition, $f$ is the data such as images and 
$u$ is the feature tensor such that
\begin{equation}
	\label{positive-u}
	u\ge 0.  
\end{equation}
Feature extraction is then viewed as an iterative procedure
(c.f. \cite{xu1992iterative}) to solve \eqref{Auf}:
\begin{equation}
	\label{iterative0}
	u^{i} = u^{i-1} + B^i\ast (f - A\ast u^{i-1}), \quad i=1:\nu.
\end{equation}
This is a typical residual correction iterative scheme for solving \eqref{Auf}, where $B^i$ is called the smoother which is also implemented as a conversational kernel with multi-channel in this work.
Using,  for example,  the special activation function $\sigma(x)={\rm
	ReLU}(x) := \max\{0,x\} \ge 0$, the above iterative process can be naturally modified to
preserve the constraint
\eqref{positive-u}: 
\begin{equation}
	\label{iterative}
	u^{i} = u^{i-1} + \sigma \circ B^i\ast \sigma  (f -  A
	\ast u^{i-1}), \quad i=1:\nu.
\end{equation}
This forms the basic block of MgNet, precisely as in~\cite{he2019mgnet}.
partial differential equations (PDEs)~\cite{xu2002method, xu2017algebraic}, 
we introduce this residual
\begin{equation}
	\label{residual}
	r^i=f-A\ast u^i.
\end{equation}
Now, the iterative process \eqref{iterative} can be written, in terms of the residual $r^i$ as:
\begin{equation}
	\label{modified-resnet}
	r^{i} = r^{i-1} - A\ast \sigma \circ B^i\ast\sigma(r^{i-1}), \quad i=1:\nu.
\end{equation}
The iterative scheme~\eqref{modified-resnet} shares an almost identical structure with the basic block architecture of pre-act ResNet. Then, the analysis process shown above will be used to understand pre-act ResNet and to develop modified ResNet and pre-act ResNet models in this paper.

Furthermore, by drawing on the multigrid~\cite{xu1992iterative,hackbusch2013multi} idea to restrict the residuals, we have a natural explanation for pooling operations in pre-act ResNet, which provides a basis for establishing a complete connection between pre-act ResNet and MgNet.
Finally, we present numerical evidence to demonstrate that our constrained linear models \eqref{Auf} and \eqref{positive-u} with the nonlinear iterative solver \eqref{iterative} or \eqref{modified-resnet} provide a second interpretation and improvement on ResNet- and MgNet-type models. The main contributions of this paper can be summarized as follows:
\begin{itemize}
	\item A constrained linear data-feature mapping is proposed and developed as an interpretable model to demonstrate the dual relation between ResNet- and MgNet-type models.
	\item Some natural modifications of ResNet-type models based on the constrained linear data-feature mapping are proposed.
	\item A general data-feature iterative scheme based on constrained linear data-feature mapping is proposed to show the rationality of MgNet.
	\item A systematic numerical study of MgNet is proposed to show its success in image classification problems and demonstrate its advantages over established networks in this context. 
\end{itemize}

This paper is organized as follows. In Section~\ref{sec:relatedwork}, 
we review some related works. In Section~\ref{sec:ResNetformula}, we introduce
precise mathematical formulas to define ResNet-type and MgNet models.
In Section~\ref{sec:constrainedmodel}, we propose the constrained linear data-feature mapping model to understand ResNet and MgNet architecture from the perspective of solving the constrained linear system based on theoretical observations and analysis. Then, we develop some modified ResNet models based on the constrained linear data-feature mapping presented. Finally, we propose a general data-feature iterative model to further demonstrate the rationality of MgNet. 
In Section~\ref{sec:numerics}, we demonstrate the validity of 
the constrained linear data-feature mapping assumption by comparing our modified ResNet-type models with the established ResNet-type models. In addition, we provide a systematic numerical study on MgNet. In Section~\ref{sec:conclusion}, we offer some concluding remarks, including a brief discussion of the implications of the results reported herein and the investigative directions that can advance this research.

\section{Related work}\label{sec:relatedwork}
In~\cite{he2019mgnet}, a unified neural network framework was proposed, known as MgNet, to establish the connections between ResNet-type CNNs and multigrid methods. In that work, the basic block of MgNet  was first introduced, as in~\eqref{iterative}.  These elementary components in block~\ref{iterative}, including the residual term $f-A\ast u$, the convolutional operators $A$ and $B^i$, the activation functions $\sigma$, and the positions of these two $\sigma$, were initially motivated by the deep connection between multigrid methods and ResNet. However, a natural interpretation of the underlying mechanism of the basic block iteration is still lacking ~\eqref{iterative}. 
Furthermore, in this paper we propose the constrained linear model  to interpret the basic block~\eqref{iterative} from the iterative method perspective.
Before MgNet, ideas and techniques from multigrid methods had been used to develop efficient CNNs.
The researchers who developed ResNet~\cite{he2016deep} first took the multigrid methods as evidence to support what is known as a residual representation for the interpretation of ResNet.
Further, \cite{ke2016multigrid, haber2017learning, zhang2019scan} adopted multi-resolution ideas to improve the performance of their networks. 
Additionally, a CNN model with a structure similar to that of the V-cycle multigrid was proposed to address volumetric medical image segmentation and biomedical image segmentation in~\cite{ronneberger2015u, milletari2016v}.  
The literature also includes studies focused on applying deep learning techniques in multigrid and numerical PDEs~\cite{katrutsa2017deep,hsieh2018learning}.

Considering the connections between CNN models and some computational mathematics methods,
researchers have also proposed the dynamic system or optimization perspective~
\cite{haber2017learning,e2017a, chang2017multi, lu2018beyond, chen2018neural}.
A key motivation of the dynamic systems viewpoint is that the iterative scheme $x^i = x^{i-1} + f(x^{i-1})$ in pre-act ResNet resembles the forward Euler scheme in numerical dynamic systems. 
Following this idea, \cite{sonoda2017double, li2017a} interpreted the data flow in ResNet as the solution of the transport equation in the characteristic line.
Furthermore, \cite{lu2018beyond} interpreted some different CNN models with residual block as some special discretization schemes for ordinary differential equations (ODEs), for example PloyNet~\cite{zhang2017polynet}, FractalNet~\cite{larsson2016fractalnet}, and RevNet~\cite{gomez2017reversible}.
Ignoring the specific discretization methods, \cite{chen2018neural} proposed a family of CNN models based on black-box solvers for ODEs. 
Some types of CNN architecture are further designed based on the iterative schemes of optimization algorithms~\cite{gregor2010learning, sun2016deep, li2018optimization}. 
These studies share the philosophy that many optimization algorithms
can be considered as discretization schemes for some special ODEs~\cite{helmke2012optimization}.

Considering the resemble properties of ResNet, \cite{veit2016residual,littwin2016loss} claim that ResNet is an ensemble of shallower models, and that discarding the intermediate residual block does not influence the model accuracy. \cite{huang2018learning,nitanda2018functional} point out that ResNet optimizes the risk in a functional space by combining an ensemble of effective features. 
	In addition, some works, such as \cite{allen2019can,huang2020deep,tirer2022kernel}, propose to study the generalization and smoothness properties of ResNet from the Neural Tangent Kernel perspective. 
	As for the approximation properties of ResNet, \cite{lin2018resnet} demonstrates that a very deep ResNet with stacked modules, that have one neuron per hidden layer, and ReLU activation functions can uniformly approximate any Lebesgue integrable function. Recently, \cite{he2022approximation} studied and proved the approximation properties of ResNet and MgNet with multi-channel $3\times3$ kernels for functions with image-type inputs, i.e. functions defined on $\mathbb R^{d\times d}$.

\section{Precise mathematical formulas for ResNet and MgNet}\label{sec:ResNetformula}
In this section, we introduce ResNet~\cite{he2016deep} and pre-act ResNet~\cite{he2016identity} with precise mathematical formulas. 
Then, we introduce MgNet~\cite{he2019mgnet} and its variants.

\subsection{ResNet and Pre-act ResNet}
\begin{figure}[H]
	\centering
	\includegraphics[width=0.8\linewidth]{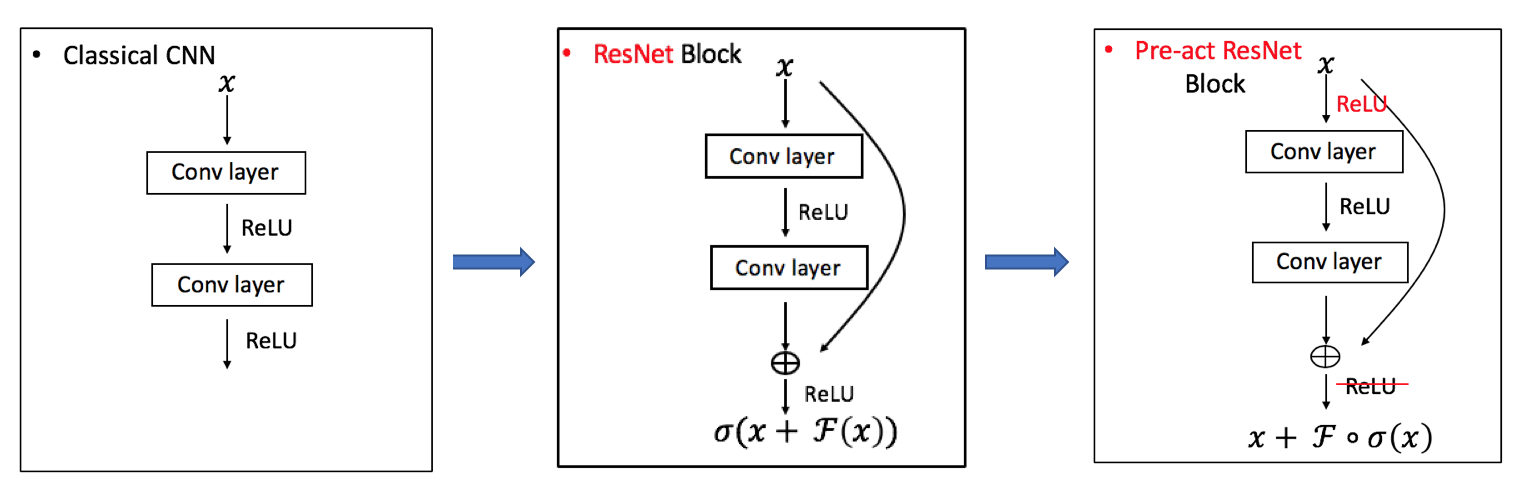}
	\caption{Comparison of classical CNN, ResNet, and pre-act ResNet.}
	\label{fig:resent}
\end{figure}

Figure~\ref{fig:resent} demonstrates the connection and 
difference between classical CNN, ResNet~\cite{he2016deep}, and pre-act ResNet~\cite{he2016identity}.
Here, $\sigma(x) = {\rm ReLU}(x) := \max\{0,x \}$ is the standard ReLU activation function.
For ResNet and pre-act ResNet with the basic block $\mathcal F(x) = A \ast \sigma \circ B \ast  x$, $A$ and $B$
are $3\times3$ convolutions with multichannel, zero padding, and stride one, and ``$\circ$'' means composition.

In order to investigate the interpretable mathematical model underlying these models, let us write these CNN models with precise mathematical formulas.
The main structure of the pre-act ResNet without the last fully connected and soft-max layers can be written as in Algorithm \ref{alg:presnet}.
\begin{algorithm}[H]
	\footnotesize
	\caption{$ h = \text{pre-act ResNet}(f; J,\nu_1, \cdots, \nu_J)$}
	\label{alg:presnet}
	\begin{algorithmic}[1]
		\STATE Initialization:  $r^{1,0} = f_{\rm in}(f)$.
		\FOR{$\ell = 1:J$}
		\FOR{$i = 1:\nu_\ell$}
		\STATE Basic Block:
		\begin{equation}\label{ori-ResNet}
			r^{\ell,i} = r^{\ell, i-1} + A^{\ell,i} \ast  \sigma \circ B^{\ell,i}\ast   \sigma (r^{\ell,i-1}).
		\end{equation}
		\ENDFOR
		\STATE Pooling(Restriction):
		\begin{equation}
			\label{ori-ResNet0}
			r^{\ell+1,0} = R_\ell^{\ell+1} \ast_2  r^{\ell, \nu_\ell} + A^{\ell+1,0} \circ \sigma \circ B^{\ell+1,0} \ast_2  \sigma (r^{\ell, \nu_\ell} ).
		\end{equation}
		\ENDFOR
		\STATE Final average pooling layer:
		$h =  R_{\rm ave}( r^{L,\nu_\ell})$.
	\end{algorithmic}
\end{algorithm}

Here, $f_{\rm in}(\cdot)$ depends on the dataset and problems 
such as $f_{\rm in}(f) = \sigma \circ \theta^0 \ast f $ for CIFAR~\cite{krizhevsky2009learning} and
$f_{\rm in}(f) = R_{\rm max}\circ \sigma \circ \theta^0 \ast  f$ for ImageNet~\cite{deng2009imagenet} as in~\cite{he2016identity}.
$r^{\ell,i} =  r^{\ell, i-1} +  A^{\ell,i} \ast  \sigma \circ B^{\ell,i} \ast \sigma (r^{i-1})$ is often called the basic ResNet block, where $A^{\ell,i}$ with $i\ge0$ and $B^{\ell,i}$ with $i\ge1$ are general $3\times3$ convolutions with zero padding and stride 1.
In the pooling block (\ref{ori-ResNet0}), $\ast _2$ means the convolution with zero padding and stride 2;
$R_\ell^{\ell+1}$ is taken as $1\times1$ kernel and referred to as the projection operator in MgNet~\cite{he2016identity};
and $B^{\ell,0}$ is taken as $3\times3$ convolutions, with the same channel dimension as the output channel dimension of $R_\ell^{\ell+1}$. 
During two consecutive pooling blocks, index $\ell$ refers to the fixed resolution or $\ell$-th level grid as in the multigrid methods. In the final average pooling layer, $R_{\rm ave}$ means average pooling whereby the stride depends on the dataset and the problem considered.

The scheme of the original ResNet~\cite{he2016deep}, which was actually developed earlier than pre-act ResNet, is very similar to that of pre-act ResNet but with a different order of convolutions 
and activation functions.
For ResNet, the basic block and pooling operations are defined by
\begin{align}
	r^{\ell,i} &= \sigma \left( r^{\ell, i-1} +  A^{\ell,i} \ast  \sigma \circ B^{\ell,i} \ast r^{\ell,i-1}\right), \label{eq:pre-actResNet1}\\
	r^{\ell+1,0} &=  \sigma \left( R_\ell^{\ell+1} \ast _2 r^{\ell, \nu_\ell} + A^{\ell+1,0} \ast  \sigma \circ B^{\ell+1,0} \ast _2r^{\ell, \nu_\ell}  \right). \label{eq:pre-actResNet2}
\end{align}

\subsection{MgNet and its variants}
In this subsection, we introduce the plain version of MgNet, and then discuss how to obtain variants of MgNet
based on choosing different hyper-parameters in the plain MgNet. 

\subsubsection{Plain MgNet structure}
Following the definitions and notations in~\cite{he2019mgnet}, we show the
plain version of MgNet in Algorithm \ref{alg:mgnet}.
\begin{algorithm}[H]
	\footnotesize
	\caption{$u^J=\text{MgNet}(f)$}
	\label{alg:mgnet}
	\begin{algorithmic}[1]
		\STATE {\bf Input}: number of grids J, number of smoothing iterations $\nu_\ell$ for $\ell=1:J$, 
		number of channels $c_{f,\ell}$ for $f^\ell$ and $c_{u,\ell}$ for $u^{\ell,i}$ on $\ell$-th grid.
		\STATE Initialization:  $f^1 = f_{\rm in}(f)$, $u^{1,0}=0$
		\FOR{$\ell = 1:J$}
		\FOR{$i = 1:\nu_\ell$}
		\STATE Feature extraction (smoothing):
		\begin{equation}\label{eq:mgnet}
			u^{\ell,i} = u^{\ell,i-1} + \sigma \circ B^{\ell,i} \ast \sigma\left({f^\ell -  A^{\ell} \ast u^{\ell,i-1}}\right) \in \mathbb{R}^{c_{u,\ell}\times n_\ell\times m_\ell}.
		\end{equation}
		\ENDFOR
		\STATE Note: 
		$
		u^\ell= u^{\ell,\nu_\ell} 
		$
		\STATE Interpolation and restriction:
		\begin{equation}
			\label{eq:interpolation}
			u^{\ell+1,0} = \Pi_\ell^{\ell+1}\ast_2u^{\ell} \in \mathbb{R}^{c_{u,\ell+1}\times n_{\ell+1}\times m_{\ell+1}} .
		\end{equation}
		\begin{equation}
			\label{eq:restrict-f}
			f^{\ell+1} = R^{\ell+1}_\ell \ast_2 (f^\ell - A^\ell \ast u^{\ell}) + A^{\ell+1} \ast u^{\ell+1,0} \in \mathbb{R}^{c_{f,\ell+1}\times n_{\ell+1}\times m_{\ell+1}} .
		\end{equation}
		\ENDFOR
	\end{algorithmic}
\end{algorithm}
Similar to ResNet, we consider $u^{\ell,i} = u^{\ell,i-1} + \sigma \circ B^{\ell,i} \ast \sigma\left({f^\ell -  A^{\ell} \ast u^{\ell,i-1}}\right) \in \mathbb{R}^{c_{u,\ell}\times n_\ell\times m_\ell}$ to be the basic MgNet block.
Here, $B^{\ell,i}$ with $i\ge1$ are general $3\times3$ convolutions with zero padding and stride 1, which 
are interpreted as the smoother convolutions in multigrid. $A^{\ell}$ is also a 
$3\times3$ convolution with zero padding and stride 1 and 
is interpreted as the system operation as in the multigrid method. A key feature of MgNet that differs from the ResNet structure is that $A^{\ell}$ does not depend on the number of iterations on each grid.
As discussed in~\cite{he2019mgnet}, this can be understood as indicating that there is only one system to be solved on each grid.
In interpolation and the restriction block (pooling block in ResNet), $\ast _2$ means convolution with zero padding and stride 2, $\Pi_\ell^{\ell+1}$ and $R^{\ell+1}_\ell$ are taken as $1\times1$ kernel. 

\subsubsection{Variants of MgNet based on different hyper-parameters}
Based on the plain MgNet in Algorithm~\ref{alg:mgnet}, 
it is natural to derive variants of MgNet by setting different hyper-parameter values. 
For simplicity, we use the following notation to represent different MgNet models with different 
hyper-parameters:
\begin{equation}\label{key}
	{\rm MgNet}[\nu_1,\cdots,\nu_J]\text{-}[(c_{u,1}, c_{f,1}), \cdots, (c_{u,J}, c_{f,J})]\text{-}B^{\ell,i}.
\end{equation}
These hyper-parameters are defined as follows:
\begin{itemize}
	\item $[\nu_1,\cdots,\nu_J]$: The number of smoothing iterations on each grid. For example, $[2,2,2,2]$ means that there are 4 grids and the number of iterations on each grid is 2.
	\item $[(c_{u,1}, c_{f,1}), \cdots, (c_{u,J}, c_{f,J})]$: The number of channels for $u^{\ell,i}$ and $f^\ell$ on each grid. We focus on the $c_{u,\ell} = c_{f,\ell}$ case, which suggests this simplification notation $[c_{1}, \cdots, c_{J}]$, or even $[c]$ if we further take $c_{1}=c_2=\cdots=c_{J}$. For example, 
	${\rm MgNet}[2,2,2,2]\text{-}[64,128,256,512]$ and ${\rm MgNet}[2,2,2,2]\text{-}[256]$.
	\item $B^{\ell,i}$: This means that we use a different smoother $B^{\ell,i}$ in each smoothing iteration. Correspondingly, $B^{\ell}$ means that we share the smoother across all the grids:
	\begin{equation}\label{eq:Bl}
		u^{\ell,i} = u^{\ell,i-1} + \sigma \circ B^{\ell} \ast \sigma\left({f^\ell -  A^{\ell} \ast u^{\ell,i-1}}\right).
	\end{equation}
	Here, we note that we always use $A^{\ell}$, which depends only on the grids.
\end{itemize}
For example, the notation ${\rm MgNet}[2,2,2,2]\text{-}[256]\text{-}B^{\ell}$
denotes an MgNet model with 4 different grids (feature resolutions), 2 smoothing iterations on each grid, 256 channels for both the feature tensor $u^{\ell,i}$ and the data tensor $f^\ell$, and \eqref{eq:Bl} as the smoothing iteration. 

\section{Constrained linear data-feature mapping}\label{sec:constrainedmodel}
In this section, we establish a new understanding of pre-act ResNet and MgNet
by drawing on the idea that the pre-act ResNet block and MgNet block are 
iterative schemes for solving some 
hidden model in each grid in a dual relation.
Then, we adopt this assumption for the ResNet-type models 
and obtain some modified models with a special parameter-sharing scheme.

\subsection{Constrained linear data-feature mapping and iterative methods}
Here, we introduce the data and feature space of CNN, which is analogous to the 
function space and its duality in the theory of multigrid 
methods~\cite{xu2017algebraic}. 
Specifically, following~\cite{he2019mgnet} we introduce 
the next data-feature mapping model in every grid level as follows:
\begin{equation}\label{eq:fmapping}
	A^{\ell} \ast u^\ell = f^{\ell},
\end{equation}
where $f^\ell$ and $u^\ell$ belong to the data and feature space of the $\ell$-th grid. 
We now make the following two important observations for this data-feature mapping:
\begin{itemize}
	\item The mapping in \eqref{eq:fmapping} is linear. More specifically, it is simply a convolution with multichannel, zero
	padding, and stride one as in pre-act ResNet or MgNet.
	\item In each level, namely between two consecutive poolings, there is only one
	data-feature mapping. Or, we say that $A^\ell$ depends only on $\ell$, but not on the number of layers.
\end{itemize}
The assumption that this linear data-feature mapping
depends only on the grid level $\ell$ is motivated from a basic property of 
multigrid methods~\cite{xu1992iterative, hackbusch2013multi, xu2017algebraic}.

In addition to \eqref{eq:fmapping}, we introduce an important constrained condition
in feature space whereby
\begin{equation}\label{eq:positive}
	u^{\ell,i} \ge 0.
\end{equation}
The rationality of this constraint in feature space can be interpreted as follows.
First, from the real neural system, the real neurons will only be
active if the electric signal is greater than a certain threshold value, i.e. human brains can only see features 
with a certain threshold.
On the other hand, the ``shift'' invariant property of feature space in CNN models,
namely, $u+a$, will not change the classification results. This means that $u+a$ should
have the same classification result with $u$. That is, we can assume
$u \ge 0$ to reduce the redundancy of $u$.

Based on these assumptions, the next step is to
solve the data-feature mapping equation in \eqref{eq:fmapping}.
We adopt some classical iterative methods~\cite{xu1992iterative} in scientific computing
to solve the system \eqref{eq:fmapping} and obtain 
\begin{equation}\label{BAmapping}
	u^{\ell,i} = u^{\ell,i-1} + B^{\ell,i} \ast (f^{\ell} - A^{\ell}\ast u^{\ell,i-1}),~ i = 1:\nu_\ell,
\end{equation}
where $u^{\ell} \approx u^{\ell,\nu_\ell}$. 
For a more detailed account of iterative methods
in numerical analysis, we refer to~\cite{xu1992iterative, hackbusch1994iterative, golub2012matrix}.
To preserve \eqref{eq:positive}, we naturally use the ReLU activation function $\sigma$
to modify \eqref{BAmapping} as follows:
\begin{equation}
	\label{eq:uBfAu}
	u^{\ell,i} = u^{\ell,i-1} + \sigma \circ B^{\ell,i}\ast \sigma  (f^\ell -  A^\ell
	\ast u^{\ell,i-1}), \quad i=1:\nu_\ell,
\end{equation}
which is exactly the same as the basic block in MgNet as in Algorithm~\ref{alg:mgnet}.

Because of the linearity of convolution in data-feature mapping,
if we consider the residual $r^{\ell,j} = f^{\ell} - A^{\ell}\ast u^{\ell,j}$, 
\eqref{eq:uBfAu} leads to the next iterative forms for the residuals:
\begin{equation}\label{eq:pre-actResNet_residual}
	r^{\ell, i} = r^{\ell,i-1} - A^\ell \ast \sigma \circ B^{\ell,i}\ast \sigma(r^{\ell,i-1}).
\end{equation}
This is the same as \eqref{eq:pre-actResNet1} under the constraint $A^{\ell,i} = A^{\ell}$ in pre-act ResNet.

We summarize the above derivation in the following theorem.
\begin{theorem}\label{thm:1} Under the assumption that there is only
	one linear data-feature mapping in each grid $\ell$, i.e., $A^{\ell,i} = A^{\ell}$, 
	the iterative form in feature space as in \eqref{BAmapping} is equivalent to 
	\eqref{eq:pre-actResNet_residual} if $A^\ell$ is invertible where $r^{\ell,i} = f^\ell - A^{\ell}\ast u^{\ell,i}$.
\end{theorem}

\subsection{Modified pre-act ResNet and ResNet}
In this subsection, we propose some modified ResNet and pre-act ResNet 
models based on the assumption of the constrained linear 
data-feature mapping underlying these models.
Although the scheme in \eqref{eq:pre-actResNet_residual} is closely
related to the original pre-act ResNet, there is a major difference between the two given that in \eqref{eq:pre-actResNet_residual} there is an extra
constraint, i.e., $A^{\ell,i} = A^{\ell}$. 
As a result, we obtain the next modified pre-act ResNet as follows:
\paragraph{Modified Pre-act ResNet (Pre-act ResNet-$A^\ell$-$B^{\ell,i}$)}
\begin{equation}\label{eq:rpre-actResNet}
	r^{\ell,i} =r^{\ell, i-1} + A^{\ell} \ast \sigma \circ B^{\ell,i} \ast \sigma (r^{\ell,i-1}).	
\end{equation}
Here, we make a small modification to the sign before $A^\ell$ in the formula because of the linearity of convolution.
As discussed, the modified pre-act ResNet model
is derived from constrained linear data-feature mapping 
by using a special iterative scheme.
Although we cannot obtain these connections in ResNet directly, 
formally we can make the modification from $A^{\ell,i}$ to $A^{\ell}$
into \eqref{ori-ResNet}
to obtain the corresponding modified ResNet models:
\paragraph{Modified ResNet (ResNet-$A^\ell$-$B^{\ell,i}$)}
\begin{equation}\label{eq:rResNet}
	r^{\ell,i} = \sigma \left( r^{\ell, i-1} +  A^{\ell} \ast \sigma \circ B^{\ell,i} \ast r^{\ell, i-1}\right).	
\end{equation}

\begin{figure}[h]
	\centering
	\includegraphics[width=0.38\linewidth]{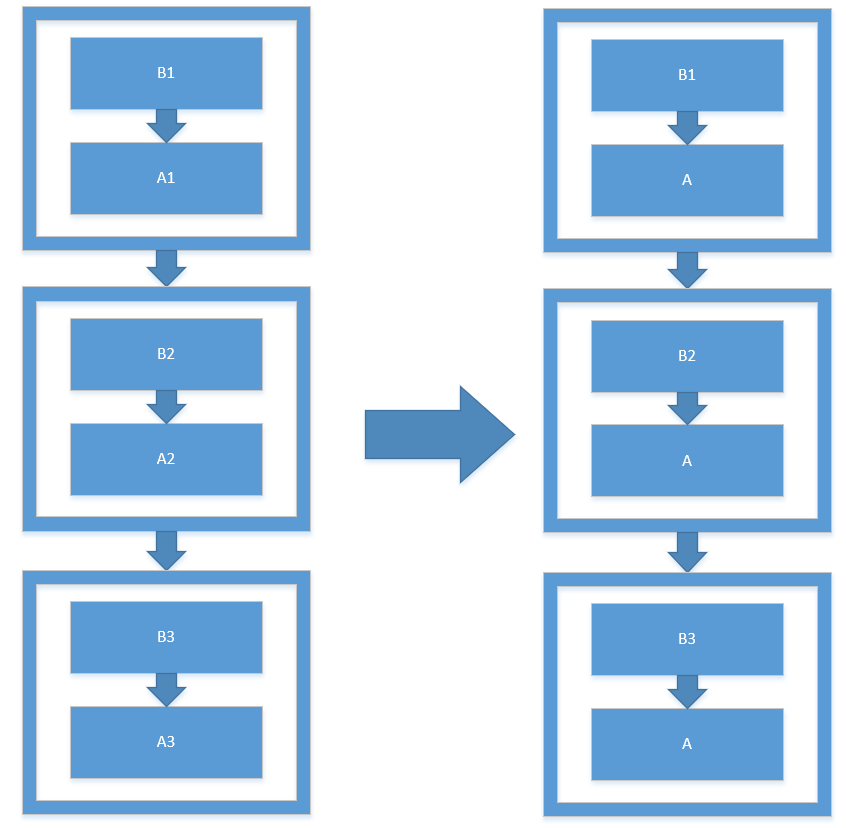}
	\caption{Diagram of modified (pre-act) ResNet basic block.}
	\label{fig.shareing}
\end{figure}

A unified but simple diagram ignoring the activation functions 
for these modified pre-act ResNet and  ResNet models with this structure is shown in Figure~\ref{fig.shareing}.

According to Theorem~\ref{thm:1}, the constrained linear model provides
a precise interpretation and understanding of the dual relation between the MgNet
model and the modified pre-act ResNet model. Briefly, MgNet applies $u^{\ell}$ as the feature
for the final logistic regression classifier. However, ResNet-type models employ $r^{\ell}$, which
is in the dual space of $u^\ell$ in multigrid theory~\cite{xu2017algebraic}.
	We provide the following three perspectives to understand the modified (pre-act) ResNet models.
	First, sharing $A$ comes as a natural result of the connections between ResNet and MgNet under the framework of the constrained linear model, since there is only one linear system on each level. However, for any given linear system $A^\ell$, applying different smoother $B^{\ell,i}$ at each residual correction step can improve the convergence of the iterative method; for example, the Chebyshev iteration and the multi-step iteration~\cite{hackbusch1994iterative,golub2012matrix} for solving linear systems.
	In addition, results in~\cite{han2015deep,he2020make} demonstrate that there is a high level of redundancy in (pre-act) ResNet models. Thus, sharing $A$ at each level (change $A^{\ell,i}$ to $A^\ell$) may not damage the representation power of (pre-act) ResNet models.
	Furthermore, we observe that $A$ and $B$ play different roles in the pre-act ResNet model in terms of their relations to the output. Recall the original basic block of pre-act ResNet
	$$
	r^{\ell,i} = r^{\ell, i-1} + A^{\ell,i} \ast \sigma \circ B^{\ell,i} \ast \sigma (r^{\ell,i-1}) = r^{\ell, i-1} + A^{\ell,i} \ast \widetilde B^{\ell,i} (r^{\ell,i-1}),
	$$
	where $\widetilde B^{\ell,i} (r) := \sigma \circ B^{\ell,i} \ast \sigma (r)$.
	Thus, we see that $r^{\ell,i}$ is linearly dependent on the kernel $A^{\ell,i}$, while $r^{\ell,i}$ is nonlinearly dependent on the kernel $B^{\ell,i}$. This simple observation indicates that sharing $B$ may have completely different effects compared with sharing $A$. Numerical results in Table~\ref{CIFAR10ANDCIFAR100} further verify this observation.
A detailed numerical study of modified ResNet-type models in the following section in Tables~\ref{MNIST} and \ref{CIFAR10ANDCIFAR100} demonstrate the rationality of the constrained linear data-feature assumption that constitutes the foundation of these modified models.

\subsection{Linear versus nonlinear data-feature mapping}\label{sec:lin-vs-nonlin}
In this subsection, we investigate the rationality of the linear assumption in data-feature mapping. We show that the linear data-feature-mapping
model is both more reasonable and more accurate than 
general data-feature mapping iterative models.

\subsubsection{A general data-feature iterative model}
Two of the most important assumptions above are that 
data-feature mapping \eqref{eq:fmapping} is a linear model and that
there should be only one model in each grid. 
To demonstrate that this linear model is adequate for image classification,
we compare it with the following nonlinear data-feature mapping:
\begin{equation}\label{eq:gfmapping}
	\mathcal A^{\ell}(u^\ell) = f^{\ell},
\end{equation}
where $\mathcal A^{\ell}$ can be chosen for some special nonlinear forms, such as 
$A^\ell \ast \sigma$, $\sigma \circ A^\ell \ast$, or $\sigma \circ A^\ell \ast \sigma$.
Then, we have the next iterative feature-extraction scheme:
\begin{equation}\label{gBAmapping}
	u^{\ell,i} = u^{\ell,i-1} + \mathcal B^{\ell,i} (f^{\ell} - \mathcal A^{\ell}(u^{\ell,i-1})),~ i = 1:\nu_\ell,
\end{equation}
where $B^{\ell,i}$ takes linear or nonlinear forms.
Here, we note that because of the nonlinearity of $\mathcal A^{\ell}$ we cannot obtain the iterative scheme for  the residuals for \eqref{gBAmapping}. 
We can execute the iteration only in the feature space. 
Thus, we propose the next general data-feature iterative model (GDFI) in Algorithm \ref{alg:GDFI}, which
follows a similar mechanism in MgNet, for example the iteration of features as in \eqref{gBAmapping} and the pooling structure as in \eqref{interpolation} and \eqref{restrict-f}.
\begin{algorithm}[H]
	\footnotesize
	\caption{$u^{J,\ell_J}=\text{ GDFI}(f; J,\nu_1, \cdots, \nu_J)$}
	\label{alg:GDFI}
	\begin{algorithmic}[1]
		\STATE Initialization:  $f^1 = f_{\rm in}(f)$, $u^{1,0}=0$
		\FOR{$\ell = 1:J$}
		\FOR{$i = 1:\nu_\ell$}
		\STATE Feature extraction (smoothing):
		\begin{equation}\label{mgnet}
			u^{\ell,i} = u^{\ell,i-1} + \mathcal B^{\ell,i}  \left({f^\ell -  \mathcal A^{\ell} (u^{\ell,i-1})}\right).
		\end{equation}
		\ENDFOR
		\STATE Pooling (interpolation and restriction):
		\begin{equation}
			\label{interpolation}
			u^{\ell+1,0} = \Pi_\ell^{\ell+1} \ast_2  u^{\ell, \nu_\ell}.
		\end{equation}
		\begin{equation}
			\label{restrict-f}
			f^{\ell+1} = R^{\ell+1}_\ell \ast_2 (f^\ell - \mathcal A^\ell(u^{\ell, \nu_\ell}) + \mathcal A^{\ell+1} (u^{\ell+1,0}).
		\end{equation}
		\ENDFOR
	\end{algorithmic}
\end{algorithm}
Here, \eqref{interpolation} and \eqref{restrict-f} are understood as different interpolation
and restriction operators because of the difference in the feature and data space as discussed in relation to MgNet. 
However, in practice, all are implemented by $3\times3$ convolution with stride 2.

\subsubsection{Numerical study indicating rationality of MgNet}
If we take this specific setting,
\begin{equation}
	\begin{aligned}
		\mathcal A^\ell(u) &= A^\ell \ast u, \\
		\mathcal B^{\ell,i}(r) &= \sigma \circ B^{\ell,i} \ast \sigma(r),
	\end{aligned}
\end{equation}
then Algorithm~\ref{alg:GDFI} precisely degenerates to MgNet. The iterative scheme for its residual, therefore, becomes
\begin{equation}
	r^{\ell, i} = r^{\ell,i-1} - A^\ell \ast \sigma \circ  B^{\ell,i}\ast \sigma(r^{\ell,i-1}),
\end{equation}
which is exactly the modified pre-act ResNet scheme discussed.

Considering the special form of MgNet, we try some numerical experiments 
with ``symmetric'' forms for different linear or nonlinear schemes for both
$\mathcal A^\ell$ and $ \mathcal B^{\ell,i}$ in Algorithm~\ref{alg:GDFI} as
\begin{equation}\label{eq:ABpossible}
	K\ast, ~K \ast \sigma,~ \sigma \circ K\ast, ~\text{and}~\sigma \circ K \ast \sigma,
\end{equation}
where $K$ is a $3\times3$ convolution kernel with multichannel, zero padding,
and stride 1.
Here, we recall that the key idea in developing pre-act ResNet~\cite{he2016identity} from ResNet~\cite{he2016deep} is to choose a better position for activation and convolution.
Thus, from another perspective, the motivation for choosing $\mathcal A^\ell$ and $ \mathcal B^{\ell,i}$ 
as in \eqref{eq:ABpossible} is to study the dual version of the idea in developing pre-act ResNet in feature space.

As the results presented in Table \ref{tab:compare-AB} show, the
original assumption about the linearity of $\mathcal A^\ell$ and the special non-linear form of $\mathcal B^{\ell,i}$, which forms MgNet exactly, is the most rational and accurate scheme. This result is
also consistent with the theoretical concern and other numerical results in the following section.

\section{Numerical experiments}\label{sec:numerics}
In this section, we design numerical experiments to show that
fixing the linear data-feature mapping in each produces only slightly
negative or sometimes even positive effects as compared with the standard ResNet and pre-act ResNet models,
which demonstrates the rationality of the constrained data-feature mapping model. 
Then, we compare the results of the MgNet model with those of established CNN models, and design a set of numerical experiments to explore the properties of MgNet and its variants.

\paragraph{Datasets.} 
We evaluate our various models using four widely used datasets: 
MNIST~\cite{lecun1998gradient}, CIFAR10~\cite{krizhevsky2009learning}, CIFAR100~\cite{krizhevsky2009learning}, and ImageNet(ILSVRC2012)~\cite{russakovsky2015imagenet}. 

\paragraph{Model implementation.}
In our experiments, the structure of the classical ResNet and pre-act ResNet models is implemented with
the same structure as in the sample codes in PyTorch~\cite{NEURIPS2019_9015}. 
We also implement our modified models and MgNet\footnote{Codes are available at \href{https://github.com/XuTeam/MgNet_Code}{https://github.com/XuTeam/MgNet\_Code.}} with PyTorch. Following the strategy in~\cite{he2016deep, he2016identity}, we adopt Batch Normalization~\cite{ioffe2015batch} 
but not Dropout~\cite{srivastava2014dropout}.

\paragraph{Training.}
We adopt the SGD training algorithm with momentum 0.9. 
We also adopt a weight decay value of 0.0005 on MNIST and CIFAR, whereas the value for ImageNet is 0.0001. 
We take the minibatch sizes to be 128, 128, 256 for  MNIST, CIFAR, and ImageNet, respectively. We use the Kaiming's weight-initialization strategy as in~\cite{he2015delving}. 
We always start training with a learning rate of 0.1. 
For MNIST, we terminate training at 60 epochs and divide the learning rate by 10 at the 50-th epoch. 
For both CIFAR and ImageNet, we terminate training at 150 epochs and divide the learning rate by 10 at every 30 epochs. 

\subsection{Numerical results for modified ResNet and pre-act ResNet}
To verify the optimality of the linear assumption of $\mathcal A^{\ell}$,
we retain the linearity assumption of $\mathcal A^{\ell}$ with the iterative method \eqref{gBAmapping}.
We, therefore, have the following iterative scheme for residuals $r^{\ell,i} = f^\ell - \mathcal A^{\ell}(u^{\ell,i})$:
\begin{equation}\label{eq:gresidual-iter1}
	r^{\ell, i} = r^{\ell,i-1} - \mathcal A^\ell \circ \mathcal B^{\ell,i}(r^{\ell,i-1}).
\end{equation}
If we take the specific setting
\begin{equation}
	\begin{aligned}
		\mathcal A^\ell (u) &= A^\ell \ast u, \\
		\mathcal B^{\ell,i}(r) &= \sigma \circ B^{\ell,i} \ast \sigma(r),
	\end{aligned}
\end{equation}
the iterative scheme for the residuals (\ref{eq:gresidual-iter1}) becomes
\begin{equation}
	r^{\ell, i} = r^{\ell,i-1} - A^\ell \ast \sigma \circ  B^{\ell,i}\ast \sigma(r^{\ell,i-1}),
\end{equation}
which is precisely the modified pre-act ResNet scheme (\ref{eq:rpre-actResNet}). We compare it with some other linear forms and nonlinear forms for both $\mathcal A^\ell$ and $ \mathcal B^{\ell,i}$ in Table \ref{tab:compare-AB}, which shows that the modified pre-act ResNet scheme (\ref{eq:rpre-actResNet}) is the most accurate. The result verifies that the assumption about the linearity of $\mathcal A^\ell$ and the special nonlinear form of $\mathcal B^{\ell,i}$ gives the most rational and accurate scheme, which is also consistent with the theoretical explanation in this paper.
\begin{table}[h]
	\begin{center}
		\begin{tabular}{lc}
			\hline
			Schemes of $\mathcal A^{\ell}$ and $\mathcal B^{\ell,i}$   & Accuracy	\\
			\hline
			$\mathcal A^{\ell} = A^{\ell} \ast$,~ $\mathcal B^{\ell,i} = B^{\ell,i}\ast $  	&71.36 \\
			$\mathcal A^{\ell} = A^{\ell} \ast$, ~$\mathcal B^{\ell,i} =  \sigma \circ B^{\ell,i} \ast$     &93.04 \\
			$\mathcal A^{\ell} = A^{\ell} \ast$, ~$\mathcal B^{\ell,i} = B^{\ell,i} \ast \sigma$     &93.80 \\
			{$\mathcal A^{\ell} = A^{\ell} \ast$, ~$\mathcal B^{\ell,i} = \sigma \circ B^{\ell,i} \ast \sigma $}     &{\bf 94.21} \\
			\hline
			$\mathcal A^{\ell} = A^{\ell} \ast \sigma $, ~$\mathcal B^{\ell,i}$ = $B^{\ell,i} \ast$     & 92.90 \\
			$\mathcal A^{\ell} = A^{\ell} \ast \sigma $, ~$\mathcal B^{\ell,i} =  \sigma \circ B^{\ell,i} \ast$     & 92.87 \\
			$\mathcal A^{\ell} = A^{\ell} \ast \sigma $, ~$\mathcal B^{\ell,i} = B^{\ell,i} \ast \sigma $   & 94.01 \\
			$\mathcal A^{\ell} = A^{\ell} \ast \sigma $, ~$\mathcal B^{\ell,i} = \sigma \circ B^{\ell,i} \ast \sigma $     & 93.98 \\
			\hline
			$\mathcal A^{\ell} = \sigma \circ A^{\ell} \ast$, ~$\mathcal B^{\ell,i}$ = $B^{\ell,i} \ast$     &92.13 \\
			$\mathcal A^{\ell} = \sigma \circ A^{\ell} \ast$, ~$\mathcal B^{\ell,i} =  \sigma \circ B^{\ell,i} \ast$    &92.44 \\
			$\mathcal A^{\ell} = \sigma \circ A^{\ell} \ast$, ~$\mathcal B^{\ell,i} = B^{\ell,i} \ast \sigma $ &93.79 \\
			$\mathcal A^{\ell} = \sigma \circ A^{\ell} \ast$, ~$\mathcal B^{\ell,i} = \sigma \circ B^{\ell,i} \ast \sigma $ 	&93.57 \\
			\hline	
			$\mathcal A^{\ell} = \sigma \circ A^{\ell} \ast \sigma$, ~$\mathcal B^{\ell,i}$ = $B^{\ell,i} \ast$   &93.20\\
			$\mathcal A^{\ell} = \sigma \circ A^{\ell} \ast \sigma$, ~$\mathcal B^{\ell,i} =  \sigma \circ B^{\ell,i} \ast$   &93.93\\
			$\mathcal A^{\ell} = \sigma \circ A^{\ell} \ast \sigma$, ~$\mathcal B^{\ell,i} = B^{\ell,i} \ast \sigma $   &94.08\\
			$\mathcal A^{\ell} = \sigma \circ A^{\ell} \ast \sigma$, ~$\mathcal B^{\ell,i} = \sigma \circ B^{\ell,i} \ast \sigma $   &94.11\\
			\hline
		\end{tabular}
	\end{center}
	\caption{Accuracy of models from Algorithm~\ref{alg:GDFI} with different linear and non-linear schemes of $\mathcal A$ and $\mathcal B$ on CIFAR10.}
	\label{tab:compare-AB}
\end{table}

Modified pre-act ResNet can also be understood as a special
parameter-sharing form on $A^{\ell,i}$.
To show that the effectiveness of the linear model does not arise from the redundancy of the CNN models, we also apply the parameter-sharing technique to $B^{\ell,i}$ for both ResNet and pre-act ResNet:
\begin{description}
	\item[Pre-act ResNet-$A^{\ell,i}$-$B^\ell$] 
	\begin{equation}
		r^{\ell,i} =r^{\ell, i-1} + A^{\ell,i} \ast \sigma \circ B^{\ell} \ast \sigma (r^{\ell,i-1}), \quad i = 1:\nu_\ell.
	\end{equation}
	\item[Pre-act ResNet-$A^\ell$-$B^\ell$ ]
	\begin{equation}
		r^{\ell,i} =r^{\ell, i-1} + A^{\ell} \ast \sigma \circ B^{\ell} \ast \sigma (r^{\ell,i-1}), \quad i = 1:\nu_\ell.
	\end{equation}
\end{description}
As $B^{\ell,0}$ will change the channel number in ResNet~\eqref{ori-ResNet0} and pre-act ResNet~\eqref{eq:pre-actResNet2}, we share only $B^{\ell,i}$ for $i = 1:\nu_\ell$.
For consistency, we denote the original ResNet and pre-act ResNet models as ResNet-$A^{\ell,i}$-$B^{\ell,i}$ and pre-act ResNet-$A^{\ell,i}$-$B^{\ell,i}$, respectively.
\begin{table}[h]
	\begin{center}
		\begin{tabular}{lcc}
			\hline
			Model & Accuracy & \# Parameters \\ \hline
			ResNet18-$A^{\ell,i}$-$B^{\ell,i}$      			& 99.49 	& 11M  \\ 
			ResNet18-$A^\ell$-$B^{\ell,i}$ 			& {\bf 99.61} 	& 8.0M   \\ 
			pre-act ResNet18-$A^{\ell,i}$-$B^{\ell,i}$      & 99.63 	& 11M  \\ 
			pre-act ResNet18-$A^\ell$-$B^{\ell,i}$	& {\bf 99.67} 	& 8.0M   \\ 
			\hline
		\end{tabular} 
	\end{center}
	\caption{Accuracy and number of parameters of ResNet-18, pre-act ResNet-18, and their modified models on MNIST.}
	\label{MNIST}
\end{table}

\begin{table}[htp!]
	\begin{center}
		\begin{tabular}{lccc}
			\hline
			Model   					&  CIFAR10 & CIFAR100 & \# Parameters\\ 
			\hline
			ResNet18-$A^{\ell,i}$-$B^{\ell,i}$       		& 94.22		& 76.08		& 11M   \\ 
			ResNet18-$A^\ell$-$B^{\ell,i}$ 					& {\bf 94.34}     & {\bf76.32}  & 8.1M	\\ 
			ResNet18-$A^{\ell,i}$-$B^\ell$ 					& 93.95 	& 74.23		& 9.7M	\\ 
			ResNet18-$A^\ell$-$B^\ell$ 						& 93.30		& 74.85 	& 6.6M \\ 
			\hline
			pre-act ResNet18-$A^{\ell,i}$-$B^{\ell,i}$      & 94.31		& 76.33	 	& 11M	\\ 
			pre-act ResNet18-$A^\ell$-$B^{\ell,i}$ 	    	& {\bf 94.54} 	& {\bf 76.43} 	& 8.1M  \\ 
			pre-act ResNet18-$A^{\ell,i}$-$B^\ell$ 			& 93.96	    & 74.45     & 9.7M	\\ 
			pre-act ResNet18-$A^\ell$-$B^\ell$ 		    	& 93.63		& 74.46     & 6.6M	\\ 
			\hline
			ResNet34-$A^{\ell,i}$-$B^{\ell,i}$       		& 94.43		& 76.31		& 21M	\\ 
			ResNet34-$A^\ell$-$B^{\ell,i}$ 					& {\bf 94.78}	    & {\bf 76.44}     & 13M	\\
			ResNet34-$A^{\ell,i}$-$B^\ell$ 					& 93.98	    & 74.48	    & 15M	\\ 
			ResNet34-$A^\ell$-$B^\ell$ 						& 93.55	    & 74.46	 	& 6.7M	\\ 
			\hline 
			pre-act ResNet34-$A^{\ell,i}$-$B^{\ell,i}$ 		& 94.70		& 77.38		& 21M	\\ 
			pre-act ResNet34-$A^\ell$-$B^{\ell,i}$ 			& {\bf 94.91}	 	& {\bf77.41}     &13M	\\ 
			pre-act ResNet34-$A^{\ell,i}$-$B^\ell$			& 94.08	    & 75.32	 	& 15M	\\ 
			pre-act ResNet34-$A^\ell$-$B^\ell$ 				& 94.01	    & 74.12		& 6.7M	\\ 
			\hline
		\end{tabular} 
	\end{center}
	\caption{Accuracy and number of parameters for ResNet, pre-act ResNet, and their variants of modified versions on CIFAR10 and CIFAR100.}
	\label{CIFAR10ANDCIFAR100}
\end{table}
Importantly, in relation to the numerical results shown in Table \ref{MNIST} and Table \ref{CIFAR10ANDCIFAR100}, the modified ResNet and pre-act ResNet models achieve almost the same
accuracy as their original respective models, whereas the other models do not.
This result indicates that the constrained data-feature mapping properly captures the mathematical insight of the ResNet-related models. 

\subsection{MgNet vs. established neural networks}
According to the discussion and numerical results in \S\ref{sec:lin-vs-nonlin},
MgNet is the most natural and accurate model under
the general data-feature iterative scheme. In the following subsections,
we present a systematic numerical study to demonstrate the success
of MgNet for image classification problems and, in this context, its advantages 
over established networks.

First, we test MgNet on the CIFAR10, CIFAR100, and ImageNet datasets and compare the results with AlexNet~\cite{krizhevsky2012imagenet}, VGG~\cite{simonyan2014very}, ResNet~\cite{he2016deep}, pre-act ResNet~\cite{he2016identity}, and WideResNet~\cite{zagoruyko2016wide}. As shown in Table \ref{tab:MgNet-CNN-Results}, MgNet achieves 96\% accuracy on CIFAR10  and  79.94\% accuracy on CIFAR100. On the ImageNet dataset, MgNet achieves 78.59\% top-1 accuracy. Compared with the selected benchmark models, MgNet, therefore, is more accurate and has fewer parameters which demonstrate
its superior effectiveness as compared with the other models. 
\begin{table}[!htp]
	\begin{center}
		\begin{minipage}[b]{0.9\linewidth}
			\begin{tabular}{cccc}
				\hline
				Dataset     					& Model                     & Accuracy   &  Parameters \\
				\hline
				\multirow{6}*{CIFAR10}          &AlexNet~\cite{krizhevsky2012imagenet}\footnote{\label{mynote}Results are reported in \href{https://reposhub.com/python/deep-learning/bearpaw-pytorch-classification.html}{https://reposhub.com/python/deep-learning/bearpaw-pytorch-classification.html}}           			& 76.22	           &   2.5M                   \\
				&VGG19~\cite{simonyan2014very} \footref{mynote}    	            & 93.56            &   20.0M                  \\
				&ResNet18~\cite{he2016deep}                   & 95.28            &   11.2M               \\
				&pre-act ResNet1001~\cite{he2016identity}         & 95.08            &   10.2M               \\
				&WideResNet$28\ast 2$~\cite{zagoruyko2016wide}       & 95.83            &   36.5M               \\
				&MgNet[2,2,2,2]-256-$B^{l}$         &  {\bf 96.00}           &   8.2M                \\
				\hline
				\multirow{6}*{CIFAR100}         &AlexNet~\cite{krizhevsky2012imagenet} \footref{mynote}                   & 43.87            &  2.5M                    \\
				&VGG19~\cite{simonyan2014very} \footref{mynote}     	            & 71.95            &  20.0M                   \\
				&ResNet18~\cite{he2016deep}                   & 77.54            &  11.2M                \\
				&preact-ResNet1001~\cite{he2016identity}            & 77.29            &  10.2M                \\
				&WideResNet$40\ast 2$~\cite{zagoruyko2016wide}       & 79.50            &  36.5M                \\
				&MgNet[2,2,2,2]-256-$B^{l}$         & {\bf 79.94}            &  8.3M                \\
				\hline
				\multirow{6}*{ImageNet}         &AlexNet~\cite{krizhevsky2012imagenet}           			& 63.30            &  60.2M                    \\
				&VGG19~\cite{simonyan2014very}      	            & 74.50            &  144.0M                \\
				&ResNet18~\cite{he2016deep}                    & 72.12            &  11.2M                \\
				&preact-ResNet200~\cite{he2016identity}           & 78.34            &  64.7M                \\
				&WideResNet$50\ast 2$~\cite{zagoruyko2016wide}       & 78.10            &  68.9M                \\
				&MgNet[3,4,6,3]-[128,256,512,1024]-$B^{l,i}$&  \bf{ 78.59}      &   71.3M      \\
				\hline
			\end{tabular}
		\end{minipage}
	\end{center}
	\label{tab:MgNet-CNN-Results}
	\caption{Accuracy of MgNet and established CNN models for widely used datasets.}
\end{table}

\subsection{MgNet with different channels}
Next, we employ two MgNet variants to demonstrate the model's scalability with respect to the number of channels. 
The first version is consistent with the typical CNN models; as the grid becomes deeper, the number of channels gradually increases. In the second version, the number of channels is the same across the grids, i.e., the number of channels does not change with the grids (resolution).
For all cases on CIFAR100 (Table \ref{tab:ablation-channel-fix}), accuracy improves simultaneously with the number of parameters.  We also found that on CIFAR100, with the same number of parameters, MgNet with fixed channels is more accurate than MgNet with increasing channels, as shown in Table \ref{tab:ablation-channel-fix}. Based on this fact, the number of channels for MgNet in relation to the CIFAR datasets is fixed in each grid in the rest of this paper.
On ImageNet, as MgNet with fixed channels has a huge number of parameters, we adopt MgNet with increasing channels in the rest of the paper. As shown in Table \ref{tab:increase-channel-imagenet}, accuracy improves simultaneously on ImageNet as the number of parameters increases.
These results show that MgNet has great potential for scalability in terms of the number of channels.

\begin{table}[H]
	\begin{center}
		
		\begin{tabular}{cccc}
			\hline
			Dataset         &   $[\nu_1,\nu_2,\cdots,\nu_J], c_\ell$,               & Accuracy & Parameters\\
			\hline
			\multirow{8}*{CIFAR100} &MgNet[2,2,2,2]-[256]-$B^{l}$                   & 79.94     & 8.3M \\		
			&MgNet[2,2,2,2]-[512]-$B^{l}$                   & 81.35     & 33.1M \\
			&MgNet[2,2,2,2]-[768]-$B^{l}$                   & 81.74     & 74.4M \\
			&MgNet[2,2,2,2]-[1024]-$B^{l}$                  & 81.89     & 132.2M \\
			\cline{2-4}
			&MgNet[2,2,2,2]-[32,64,128,256]-$B^{l}$         & 74.95     & 2.3M\\			
			&MgNet[2,2,2,2]-[64,128,256,512]-$B^{l}$        & 78.06     & 12.5M\\
			&MgNet[2,2,2,2]-[128,256,512,1024]-$B^{l}$      & 80.29     & 37.5M \\
			&MgNet[2,2,2,2]-[256,512,1024,2048]-$B^{l}$     & 81.49     & 150.0M \\
			\hline			
		\end{tabular}
\end{center}
\caption{MgNet with fixed and increasing channels on CIFAR100}
\label{tab:ablation-channel-fix}
\end{table}

\begin{table}[H]
\begin{center}
	\begin{tabular}{cccc}
		\hline
		Dataset                   &  $[\nu_1,\nu_2,\cdots,\nu_J], c_\ell$,  &  Accuracy   & Parameters \\
		\hline
		\multirow{2}*{ImageNet}
		&MgNet[2,2,2,2]-[64,128,256,512]-$B^{\ell}$            &  72.32        & 9.9M             \\
		&MgNet[2,2,2,2]-[128,256,512,1024]-$B^{\ell}$          &  76.82         & 38.5M             \\
		\hline
	\end{tabular}
	
\end{center}
\caption{MgNet with increasing number of channels on ImageNet.}
\label{tab:increase-channel-imagenet}
\end{table}

\subsection{MgNet with different number of iterations $\nu_\ell$}
In this subsection, we explore the impact of the number of iterations $\nu_\ell$ of each grid in MgNet.
We change only $\nu_\ell$ in the grids and keep all the other parameters fixed.
In Table \ref{tab:nu-1-cifar100}, we can see that as $\nu_1$ increases, the corresponding accuracy improves.
We also perform similar tests on the other layers, except for the first grid; increasing the number of iterations of the other grids, $\nu_2, \nu_3, \nu_4$, has no significant impact on the accuracy of the model, as shown in Table \ref{tab:nu-234-cifar100}.
In addition, we test different $\nu_3$ values on ImageNet and find that as $\nu_3$ increases the corresponding accuracy improves, as shown in Table \ref{tab:nu-3-imagenet}.
Increasing the number of iterations $\nu_\ell$ does not increase the number of parameters, which is also an advantage that MgNet has over the other models tested.
\begin{table}[H]
\begin{center}
	\begin{tabular}{cccc}
		\hline
		Dataset                   & $[\nu_1,\nu_2,\cdots,\nu_J], c_\ell$,  &  Accuracy  & Parameters \\
		\hline
		\multirow{12}*{CIFAR100}  &MgNet[2,2,2,2]-[256]-$B^{\ell}$         &  79.94                & 8.3M             \\
		&MgNet[4,2,2,2]-[256]-$B^{\ell}$         &  80.25                & 8.3M             \\
		&MgNet[8,2,2,2]-[256]-$B^{\ell}$         &  80.32                & 8.3M             \\
		&MgNet[16,2,2,2]-[256]-$B^{\ell}$        &  80.42                & 8.3M             \\
		&MgNet[32,2,2,2]-[256]-$B^{\ell}$        &  80.89                & 8.3M             \\
		\cline{2-4}
		&MgNet[2,2,2,2]-[512]-$B^{\ell}$         &   81.35               & 33.1M            \\
		&MgNet[4,2,2,2]-[512]-$B^{\ell}$         &   81.53               & 33.1M            \\
		&MgNet[8,2,2,2]-[512]-$B^{\ell}$         &   81.83               & 33.1M            \\
		&MgNet[16,2,2,2]-[512]-$B^{\ell}$        &   81.97               & 33.1M            \\
		\cline{2-4}
		&MgNet[2,2,2,2]-[1024]-$B^{\ell}$        &   81.89               & 132.2M            \\
		&MgNet[8,2,2,2]-[1024]-$B^{\ell}$        &   82.46               & 132.2M            \\
		\hline
	\end{tabular}
\end{center}
\caption{MgNet with different $\nu_1$ on CIFAR100.}
\label{tab:nu-1-cifar100}
\end{table}

\begin{table}[H]
\begin{center}
		\begin{tabular}{cccc}
			\hline
			Dataset &                $[\nu_1,\nu_2,\cdots,\nu_J], c_\ell$   &  Accuracy  & Parameters \\
			\hline
			\multirow{12}*{CIFAR100}  &MgNet[2,2,2,2]-[256]-$B^{\ell}$                        &  79.94     &     8.3M         \\
			&MgNet[2,4,2,2]-[256]-$B^{\ell}$                        &  79.96     &     8.3M         \\
			&MgNet[2,8,2,2]-[256]-$B^{\ell}$                        &  79.92     &     8.3M         \\
			&MgNet[2,16,2,2]-[256]-$B^{\ell}$                       &  79.97     &     8.3M         \\
			\cline{2-4}
			&MgNet[2,2,2,2]-[256]-$B^{\ell}$                         &  79.94                &     8.3M         \\
			&MgNet[2,2,4,2]-[256]-$B^{\ell}$                         &  79.85                &     8.3M         \\
			&MgNet[2,2,8,2]-[256]-$B^{\ell}$                         &  79.91                &     8.3M         \\
			&MgNet[2,2,16,2]-[256]-$B^{\ell}$                        &  79.77                &     8.3M         \\
			\cline{2-4}
			&MgNet[2,2,2,2]-[256]-$B^{\ell}$                         &  79.94                &     8.3M         \\
			&MgNet[2,2,2,4]-[256]-$B^{\ell}$                         &  79.60                &     8.3M         \\
			&MgNet[2,2,2,8]-[256]-$B^{\ell}$                         &  79.28                &     8.3M         \\
			&MgNet[2,2,2,16]-[256]-$B^{\ell}$                        &  79.47                &     8.3M         \\
			\hline
		\end{tabular}
	\end{center}
	\caption{MgNet with different $\nu_2, \nu_3, \nu_4$ on CIFAR100.}
	\label{tab:nu-234-cifar100}
\end{table}

\begin{table}[H]
	\begin{center}
		\begin{tabular}{cccc}
			\hline
			Dataset                   & $[\nu_1,\nu_2,\cdots,\nu_J], c_\ell$      &  Accuracy   & Parameters \\
			\hline
			\multirow{4}*{ImageNet}  &MgNet[2,2,2,2]-[64,128,256,512]-$B^{\ell}$  &  72.32      &     9.9M         \\
			&MgNet[2,2,4,2]-[64,128,256,512]-$B^{\ell}$  &  73.04      &     9.9M         \\
			&MgNet[2,2,8,2]-[64,128,256,512]-$B^{\ell}$  &  73.72      &     9.9M         \\
			&MgNet[2,2,16,2]-[64,128,256,512]-$B^{\ell}$ &  73.81      &     9.9M         \\
			\hline
		\end{tabular}
	\end{center}
	\caption{MgNet with different $\nu_3$ on ImageNet.}
	\label{tab:nu-3-imagenet}
\end{table}

\subsection{Parameter sharing on operator $B$}
Here, we explore the parameter-sharing technique on operator $B$. We consider two cases: $B^{\ell}$ ($B$ operator of every iteration step in the same grid is the same) and $B^{\ell,i}$ ($B$ operators of every iteration step in the same grid are different). The influence of $B$ is tested on CIFAR100 and ImageNet.
As shown in Table \ref{tab:share-B-cifar100} and Table \ref{tab:share-B-imagenet}, the MgNet-$B^{\ell,i}$ models have a larger number of parameters and are more accurate compared with the MgNet-$B^{\ell}$ models.
\begin{table}[!htbp]
	\begin{center}
		\begin{tabular}{cccc}
			\hline
			Dataset                 &$[\nu_1,\nu_2,\cdots,\nu_J], c_\ell$           &  Accuracy             &Parameters \\
			\hline
			\multirow{6}*{CIFAR100} &MgNet[2,2,2,2]-[256]-$B^{\ell}$                &  79.94                &     8.3 M       \\
			&MgNet[2,2,2,2]-[256]-$B^{\ell,i}$              &  80.12                &     10.7M        \\
			&MgNet[4,2,2,2]-[256]-$B^{\ell}$                &  80.25                &     8.3 M       \\
			&MgNet[4,2,2,2]-[256]-$B^{\ell,i}$              &  80.63                &     11.9M        \\
			&MgNet[8,2,2,2]-[256]-$B^{\ell}$                &  80.32                &     8.3M         \\
			&MgNet[8,2,2,2]-[256]-$B^{\ell,i}$              &  81.42                &     14.3M        \\
			\hline
		\end{tabular}
	\end{center}
	\caption{Comparison of MgNet with share $B$ and no share $B$ on CIFAR100.}
	\label{tab:share-B-cifar100}
\end{table}

\begin{table}[H]
	\begin{center}
		\begin{tabular}{cccc}
			\hline
			Dataset                  &$[\nu_1,\nu_2,\cdots,\nu_J], c_\ell$   &  Accuracy  &Parameters \\
			\hline
			\multirow{8}*{ImageNet} &MgNet[2,2,2,2]-[64,128,256,512]-$B^{\ell}$     &  72.32         &     9.9M         \\
			&MgNet[2,2,2,2]-[64,128,256,512]-$B^{\ell,i}$   &  73.36         &     13.0M        \\
			\cline{2-4}
			&MgNet[2,2,4,2]-[64,128,256,512]-$B^{\ell}$     &  73.04         &     9.9M         \\
			&MgNet[2,2,4,2]-[64,128,256,512]-$B^{\ell,i}$   &  74.58         &     14.7M        \\
			\cline{2-4}
			&MgNet[2,2,2,2]-[128,256,512,1024]-$B^{\ell}$   &  76.82        &     38.5M        \\
			&MgNet[2,2,2,2]-[128,256,512,1024]-$B^{\ell,i}$ &  77.27        &     51.1M        \\
			\cline{2-4}
			&MgNet[2,2,4,2]-[128,256,512,1024]-$B^{\ell}$    &  77.58        &     38.5M         \\
			&MgNet[2,2,4,2]-[128,256,512,1024]-$B^{\ell,i}$  &  77.94        &     55.7M        \\
			\hline
		\end{tabular}
	\end{center}
	\caption{Comparison of MgNet with share $B$ and no share $B$ on ImageNet.}
	\label{tab:share-B-imagenet}
\end{table}

\section{Discussion and conclusion}\label{sec:conclusion}
We proposed a constrained linear data-feature-mapping model 
as underlying ResNet and MgNet to demonstrate their dual relation. 
Under this model, we investigated the connections between the traditional
iterative method with a nonlinear constraint and the basic block scheme in the pre-act ResNet model, 
and developed an explanation for pre-act ResNet at a technical level from the dual perspective of MgNet. 
In comparison with existing studies that discuss the
connection between dynamic systems and ResNet, the constrained data-feature-mapping model goes beyond both formal and
qualitative comparisons to identify key model components via a more detailed account.  
Furthermore, we hope that the reason, and the ways in which, ResNet-type models work can be
mathematically understood in a similar fashion as is the case for classical 
iterative methods in scientific computing for which the theoretical understanding is more mature and better-developed.  
The numerical experiments verified in this paper indicate the rationality and efficiency
for the constrained learning data-feature-mapping model. In addition, a systematic numerical study on MgNet shows its success in image classification problems and its advantages over established networks. 

We believe that our investigation into the connections between CNNs and classical iterative methods opens a new door to the mathematical understanding and analysis of CNN models with certain structures as well as creating opportunities to make improvements to them.  
The results presented indicate the great potential of this model from both
theoretical and empirical viewpoints.  
Obviously many aspects of classical iterative methods with constraint
should be further explored with the goal of making significant improvements in this regard. 
For example, we are currently focusing on establishing the connection of ResNet with bottleneck and the subspace correction iterative methods~\cite{xu1992iterative} and applying different techniques from iterative methods to MgNet.

\section*{Acknowledgements}
This work was partially supported by the Center for Computational Mathematics and Applications (CCMA) at The Pennsylvania State University, the Verne M. William Professorship Fund from The Pennsylvania State University, and the National Science Foundation (Grant No. DMS-1819157 and DMS-2111387). The authors thank Huang Huang for his help with partial numerical experiments.





  


\bibliographystyle{plain}
\bibliography{MgNet_Linear}
\end{document}